\pdfoutput=1

\documentclass[11pt]{article}

\usepackage[final]{acl}

\usepackage{times}
\usepackage{latexsym}

\usepackage[T1]{fontenc}

\usepackage[utf8]{inputenc}

\usepackage{microtype}

\usepackage{inconsolata}

\usepackage{graphicx}
\usepackage{subfigure}
\usepackage{graphicx}
\usepackage{amsmath}

\title{When Compression Meets Model Compression: Memory-Efficient Double Compression for Large Language Models}

\author{Weilan Wang\textsuperscript{1}   Yu Mao\textsuperscript{1}  Dongdong Tang\textsuperscript{1}
\\ {\bf Hongchao Du\textsuperscript{1}   Nan Guan\textsuperscript{1}   Chun Jason Xue}\textsuperscript{2} \\
\textsuperscript{1}City University Of Hong Kong,
\textsuperscript{2}Mohamed bin Zayed University of Artificial Intelligence
\\
\small{
\textbf{Correspondence:} {Yu Mao:yumao7-c@my.cityu.edu.hk } {Dongdong Tang:dtang8-c@my.cityu.edu.hk}
}
}

\begin{document}
\maketitle
\begin{abstract}
Large language models (LLMs) exhibit excellent performance in various tasks. However, the memory requirements of LLMs present a great challenge when deploying on memory-limited devices, even for quantized LLMs. This paper introduces a framework to compress LLM after quantization further, achieving about 2.2x compression ratio.
A compression-aware quantization is first proposed to enhance model weight compressibility by re-scaling the model parameters before quantization, followed by a pruning method to improve further. Upon this, we notice that decompression can be a bottleneck during practical scenarios. We then give a detailed analysis of the trade-off between memory usage and latency brought by the proposed method. A speed-adaptive method is proposed to overcome it. The experimental results show inference with the compressed model can achieve a 40\% reduction in memory size with negligible loss in accuracy and inference speed. 
\end{abstract}

\section{Introduction}
\vspace{-0.1in}

Large language models (LLMs) exhibit excellent performance in various natural language tasks and have made significant advancements in recent years\cite{opt,llama,zerooffload,palm}. However, LLMs with billion-scale parameters present a high demand for memory\cite{inferencesurveyllm}.
Deploying these LLMs on memory-limited devices poses a huge challenge due to their extensive memory requirements. The model size of large language models has been developing at a fast pace\cite{llminflash}\cite{flexgen}. The memory or GPU memory is far from sufficient to run LLMs, even for lightweight models. 


Model compression is one promising approach to address this challenge, referring to the methods that reduce model size\cite{zhu2023survey}\cite{wang2024modelsurvey}. 
Quantization is one of the model compression techniques widely used for compressing LLMs. It approximates the model's weights with shorter bits to reduce the model size. 
Nonetheless, the INT8 quantized 7B model (a relatively small large-language model) still needs over 7GB of memory.
A substantial gap remains due to the conflict between large model parameter sizes and insufficient memory.


This paper aims to reduce the memory gap by further compressing the quantized models. 
Recent work demonstrates that there's still compressibility in quantized LLM models\cite{mao2024compressibility}. Following this, we first employ a lossless compression algorithm\cite{Zstd} to explore the compressibility of quantized model data. 
To investigate and improve the compressibility, we analyze the data distribution before and after quantization.
The quantized data with higher compressibility presents two key observations that can be utilized for further enhanced compressibility: 1) Data with uneven distribution exhibits higher compressibility; 2) The quantized data have a higher proportion of near-zero values.

These observations suggest that modifying the data distribution can improve data compressibility. Simply applying a scaling technique to model weights makes the quantized weight distribution more uneven, which can improve compressibility but reduce accuracy by approximately 60\%.
To address this issue, we propose a compression-aware quantization and pruning approach that expands important values and reduces unimportant weight.

Although the compressed model significantly reduces memory usage for LLMs, the frequent decompression operation may also introduce overhead during the model inference process. 
%
To alleviate this problem, we first implement a compressed model inference framework and analyze overall inference speed based on system memory architecture. We then propose a speed-adaptive method to address the bottleneck caused by decompression overhead. By partially compressing the model data, we can improve the total decompress throughput. Additionally, we provide strategies to balance memory usage and inference speed.


The experimental results show a remarkable Compression Ratio (CR), about 2.2, in quantized model size. Importantly, this CR is achieved with minimal compromise on model accuracy within a 1\% drop. Inference with compressed LLM can reduce 40\% memory size without affecting the speed. 
Our approaches present a promising solution to the memory challenges faced by LLMs, enabling their deployment on memory-limited devices.

\begin{figure}
	\centering
	\subfigure[Activation]{
	       \begin{minipage}[b]{0.5\textwidth}
			\includegraphics[width=0.45\textwidth]{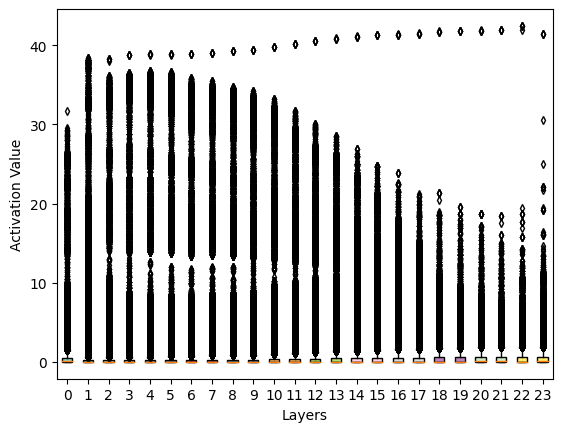} 
                \includegraphics[width=0.45\textwidth]{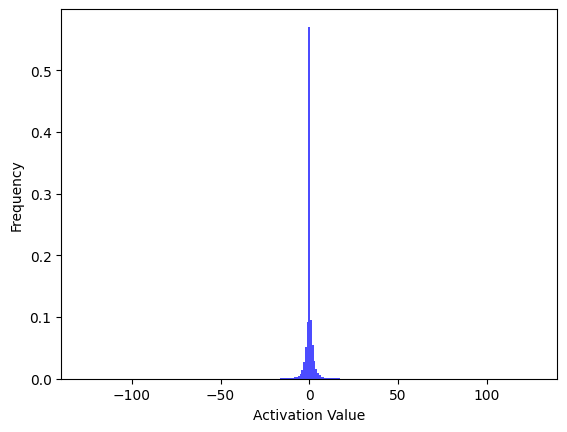}
	       \end{minipage}
	       \label{fig:act_dist}
	}
        \subfigure[Weight]{
    	\begin{minipage}[b]{0.5\textwidth}
   		   \includegraphics[width=0.45\textwidth]{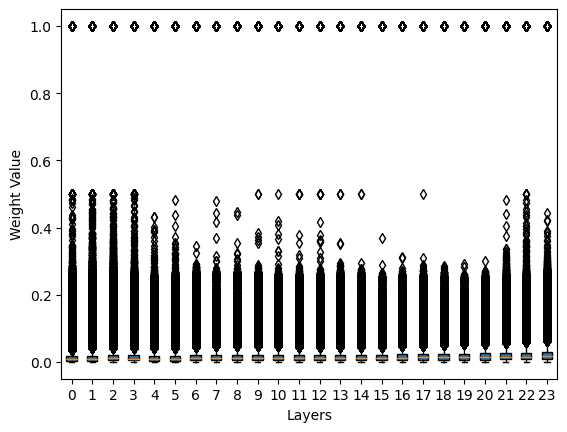}
	           \includegraphics[width=0.45\textwidth]{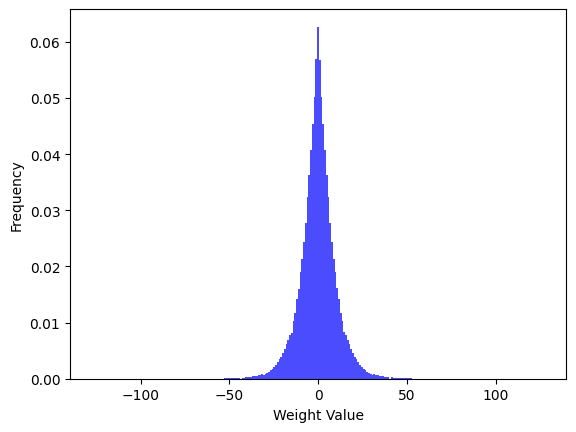}
    	\end{minipage}
		\label{fig:weight_dist}
        }
        \subfigure[Scaled-Weight]{
    	\begin{minipage}[b]{0.5\textwidth}
   		   \includegraphics[width=0.45\textwidth]{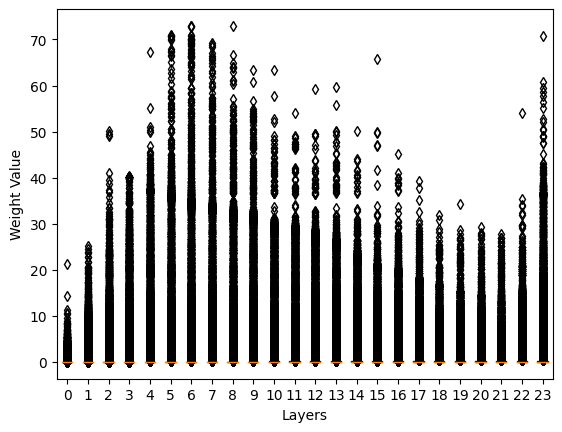}
	           \includegraphics[width=0.45\textwidth]{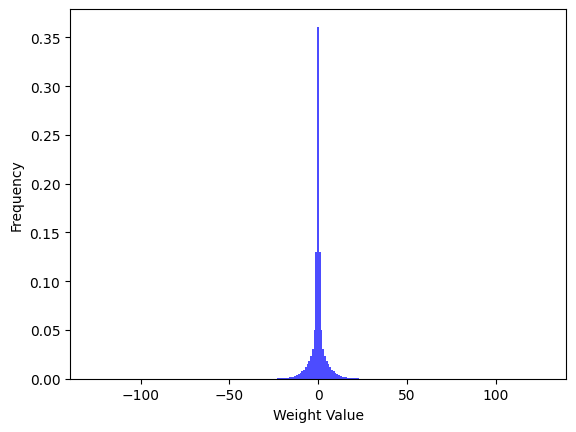}
    	\end{minipage}
		\label{fig:weight_scale_dist}
        }        
	\caption{Activation, weight and scaled-weight data distribution of OPT-1.3B Model. The left displays the data distribution for every layer before quantization. Points above the upper edge lines are outliers. The data distribution after quantization is on the right. 
 }
	\label{fig:motivation_distribution}
\end{figure}

\section{Explore the compressibility of LLMs}

\begin{table*}
  \centering
  \renewcommand{\arraystretch}{0.9}
  \begin{tabular}{llllll}
    \hline
     & \multicolumn{3}{c}{\textbf{Compression Ratio}}& \multicolumn{2}{c}{\textbf{Accuracy}}\\
    \hline
     & \textbf{IINT8 Activation}& \textbf{INT8 Weight}& \textbf{Weight-scaled}& \textbf{W8A8}&\textbf{W8A8-scaled}\\
    \hline
    OPT-1.3B & 4.30&                            1.54& 2.46& 0.57&0.32\\
    OPT-2.7B & 4.35&                            1.38& 2.59& 0.64&0.25\\
    OPT-13B & 4.98&                            1.34& 2.30& 0.65&0.23\\
    \hline
  \end{tabular}
  \caption{\label{tab: compressibility}
   Model Compression Ration and Accuracy. The CR of activation is much higher than the  CR of weight. Scaling weight can improve the CR, but lead to a huge loss of accuracy for models.
  }
\end{table*}

We first examine the compressibility characteristics of model data, focusing on model weights and activations. Our analysis reveals that temporary activations achieve a high compression ratio, whereas weights exhibit a low compression ratio. We then explore enhancing weight compressibility through scaling operations.




\noindent\textbf{Compressibility of  LLMs.}
INT8 quantized model can be further compressed.
To investigate this compressibility, we employ a general compression algorithm Zstd \cite{Zstd} to compress the quantized activations/weights data from three models of varying sizes. The compressibility is quantified using the compression ratio (CR), calculated by dividing the data size before compression by the size after compression. 

Compressing weights presents more challenges than compressing activations. However, optimizing weight compression is crucial for efficient memory use since it accounts for most of the memory overhead. As shown in Table \ref{tab: compressibility}, both activation and weight demonstrate compressibility after quantization. The CR for activations is typically much higher, often surpassing 4.0, whereas the CR for weights remains lower. Among the three evaluated models, the opt-1.3B model achieves the highest CR at 1.54, leading to a 35\% reduction in memory usage. 



\noindent\textbf{Analysis for Data Distribution.}
There is a notable gap in CR between activations and weights. To understand the reasons behind this gap, we analyze the data distribution of one specific model (OPT-1.3B). Figure \ref{fig:motivation_distribution} illustrates the data distribution for both the absolute values of FP16 data and the corresponding INT8 values after quantization.

The  FP16 activation and weight show different data distributions. The activation data range is $40\times$ larger than the weight, and there are more outliers in it. The distribution of the original FP16 activation data spans from 0 up to 40, with a significant number of outliers present in the data, and the majority of values are concentrated in the range of 0 to 1. In contrast, the weights exhibit a different pattern, with all values being smaller than 1 and most outliers falling below 0.6. The distinct distributions reveal the reason for different compressibility.


The connection between the data distribution and compressibility of the quantized model can be concluded as two key points: 1) The uneven distribution of data before quantization leads to higher compressibility. If the data distribution can be converted to an uneven distribution by introducing more outliers, the compressibility may be improved.   2) There are about 58\% near-zero values in the quantized activation, while the percentage of weight is only 6\%. Improving the zero numbers of quantized data may improve the compressibility further.

We first explored the possibility of improving weight compressibility by transforming the weight distribution into an uneven distribution. It is achieved by scaling the weight, thereby introducing more outliers in the weight distribution.
By scaling the weight distribution to mirror that of the activations, as depicted in Figure \ref{fig:weight_scale_dist}, the data range was expanded from 0 up to 70, resulting in an increased number of outliers. Subsequent quantization yielded a data distribution similar to that of the activations, including 36\% near-zero values, resulting in a 72\% improvement in compression ratio. This improvement was observed across models of different sizes.

Although scaling weights can effectively improve weight compressibility, it significantly hurts the model's accuracy. We evaluated the accuracy of the W8A8-scaled model and found a significant drop by 57\%. These findings highlight the challenges associated with achieving both weight compressibility and model accuracy.  

In this paper, we propose a novel approach to address this challenge by introducing weight scaling using activation magnitudes and increasing the occurrence of zero values. These techniques significantly enhance the compression ratio without compromising model accuracy.

\section{LLM Double Compression}
\subsection{Overview}
\begin{figure}[t]
  \includegraphics[width=\columnwidth]{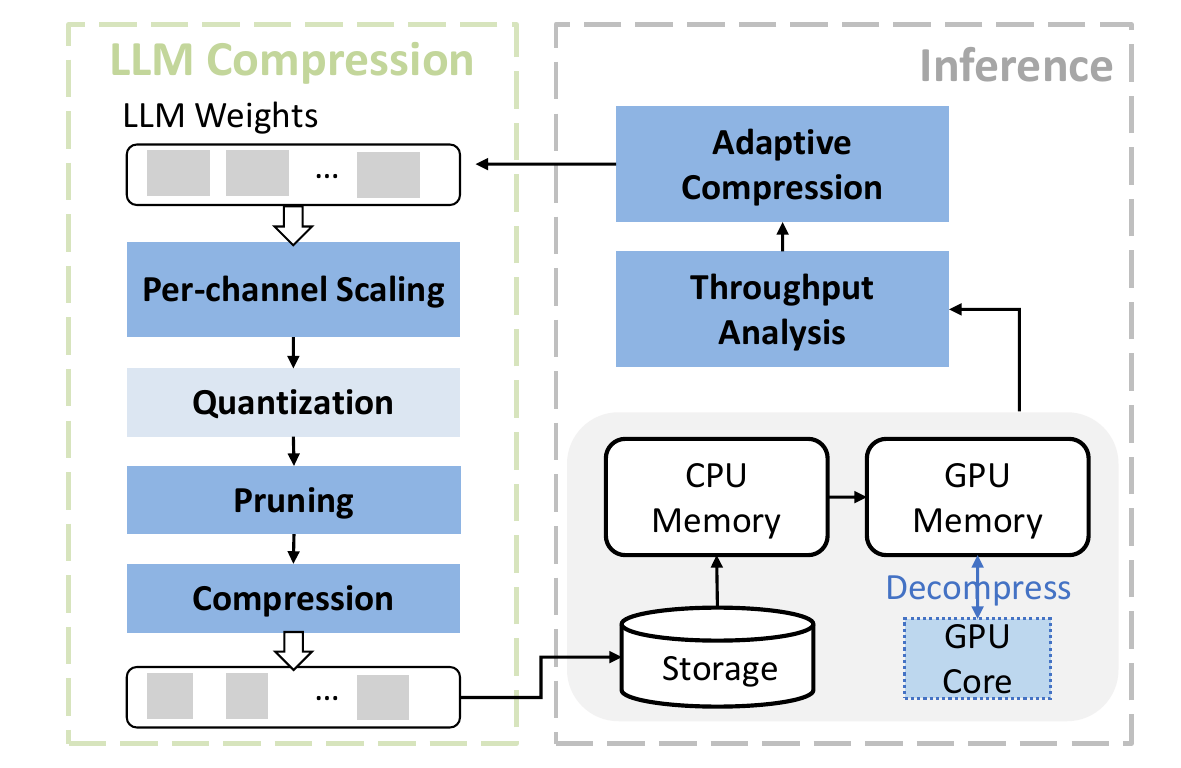}
  \caption{Overview of Double Compression. The LLM weights are scaled, quantized, pruned, and compressed. The inference throughput is analyzed for adaptive compression of model weights.}
  \label{fig:overview}
\end{figure}

The memory-efficient double compression design follows a two-stage approach, including LLM compression and runtime inference stages, as shown in Figure~\ref{fig:overview}.


To optimize the compressibility of model weights, a per-channel scaling technique is implemented prior to quantization, maximizing the compression ratio. The quantized model then undergoes pruning to increase zero-valued weights, further improving compression. Finally, a lossless algorithm compresses the weights, stored in binary.

Potential decompression overhead is investigated for compressed model inference. The compressed model is loaded to GPU memory, and a GPU decompression process decompresses weights for the current inference. Throughput analysis evaluates the impact of decompression on inference speed, leading to a speed-adaptive compression technique that selectively compresses data to reduce decompression latency.



Overall, our goal is to bridge the gap between memory usage and the size of LLMs. Combining all of the compression methods, the compressibility of LLMs is optimized without impacting the model accuracy. By employing efficient decompression strategies, double compression can be effectively utilized while minimizing the impact on inference speed.

\subsection{Improve the Compressibility of LLMs}
\begin{figure}[t]
  \includegraphics[width=\columnwidth]{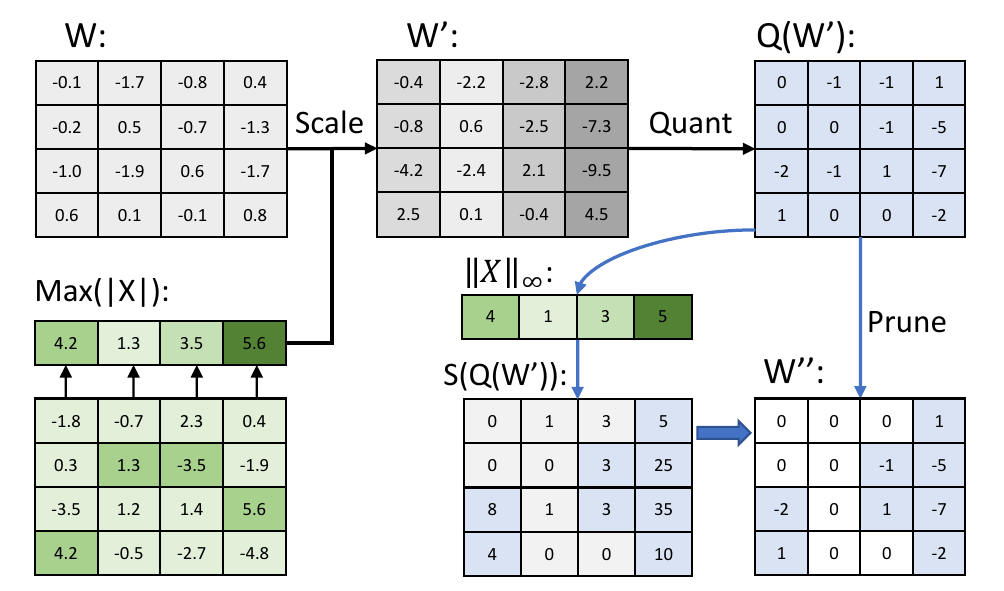}
  \caption{The double compression method first scales the weight using per-channel activation maximum value. Then, INT 8 quantization is applied to compress the weight, followed by pruning using score.}
  \label{fig:double compression}
\end{figure}

\textbf{Per-channel Scaling.}
LLMs exhibit channel-wise activation patterns \cite{smoothquant}\cite{os+}, where weight channels with higher activation magnitudes are generally more significant. Our initial proposal introduces a per-channel scaling technique for weights, with the scaling factor being optimized using model accuracy. A weight scaling example is shown in Figure \ref{fig:double compression}.


Different channels in LLM's weight capture distinct features. If we migrate the outlier from activation to weight, we can scale the weight to uneven distribution without impacting the inference result. We propose utilizing channel features in activation to enhance weight compressibility. This involves multiplying separate scaling factors with the weights of individual channels. 
For every weight, the channel is $i$, and the corresponding activation is $X_i$. 
The scale operation can be expressed as:
\begin{equation}
Y = \left(\frac{X_i}{s}\right) \times (s \times W) = X' \times W'
\label{equ:equation1}
\end{equation}
\begin{equation}
s = \max \left\lvert (X_i) \right\rvert^{\alpha}, \alpha \in \{0, 0.1, \ldots, 1\}
\label{equ:equation2}
\end{equation}

As outliers introduce quantization errors that may compromise the model's accuracy \cite{smoothquant}. The parameter $\alpha$ is utilized as a tuning parameter to control the reduction of outliers. The value of $\alpha$ may vary depending on the characteristics of each model and its desired level of accuracy.

Through per-channel scaling, we amplify the importance of weight channels that contribute significantly to the model's output, while reducing the influence of less significant channels. 
INT8 quantization is taken to the scaled weights and activations. The quantization process can be expressed as:
\begin{equation}
Y_{\text{INT8}} = \text{Quant}(X') \times \text{Quant}(W')
\label{equ:equation4}
\end{equation}
Due to the improved maximum absolute value of the weights ($\max(|W|)$) and the presence of an increased number of outliers in the weight distribution, the quantization process leads to an uneven distribution of the weight data. Consequently, the unevenly distributed data exhibits a higher degree of compressibility.

\noindent\textbf{Pruning.}
Unlike previous pruning methods\cite{wanda,sparsegpt}, we find that the l-infinite norm of activation better measures weight importance.
The l-infinite norm of activation ($\max\left(|X_i|\right)$) is used to increase the number of zero values in quantized weights. It utilizes the channel feature we get in scale operation. Compared with the start-of-art pruning method \cite{wanda}, the computational complexity $O(d^2)$ is reduced to $O(d)$.
The weight importance score is calculated using Equation \ref{equ:equation5}, and the weights with a lower score than the threshold will be set to zero. Subsequently, the model is compressed using the lossless compression algorithm. 

\begin{equation}
S(Q(W')) = \| X \|_{\infty} \cdot Q(W')
\label{equ:equation5}
\end{equation}

It is important to highlight a key distinction between our method and traditional pruning approaches. Conventional pruning stores non-zero values and an index matrix, requiring weight reconstruction and incurring computational overhead during inference. 
In contrast, our approach employs a lossless compression method to compress the model data. By leveraging the compression algorithm's capabilities, non-zero and zero values are both efficiently compressed.
This scheme effectively increases the overall CR of the model.

\subsection{Inference with DC-Compressed LLMs}

\begin{figure*}
	\centering
	\subfigure[Uncompressed LLM in GPU Memory.]{
		\begin{minipage}[b]{0.22\textwidth}
			\includegraphics[width=1\textwidth]{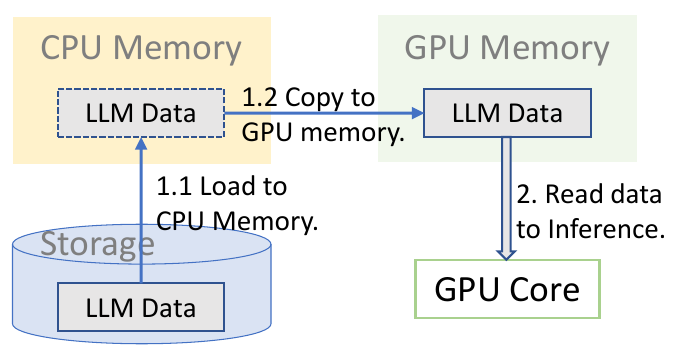}
		\end{minipage}
		\label{fig:design1}
	}
        \subfigure[Compressed LLM in GPU Memory.]{
    	\begin{minipage}[b]{0.22\textwidth}
   		  \includegraphics[width=1\textwidth]{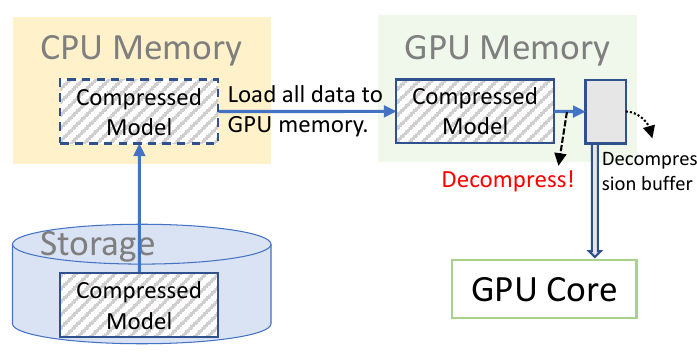}
    	\end{minipage}
		\label{fig:design2}
        }
	\subfigure[Compressed LLM in GPU-CPU Memory.]{
		\begin{minipage}[b]{0.22\textwidth}
			\includegraphics[width=1\textwidth]{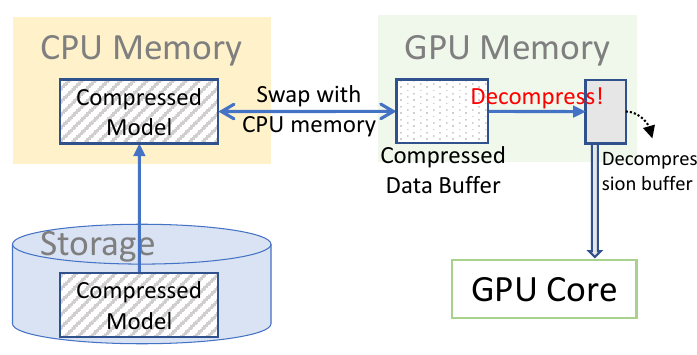}
		\end{minipage}
		\label{fig:design3}
	}
	\subfigure[Compressed LLM in Storage.]{
		\begin{minipage}[b]{0.22\textwidth}
			\includegraphics[width=1\textwidth]{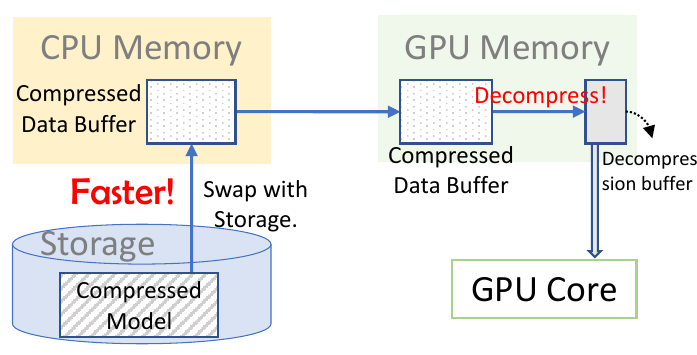}
		\end{minipage}
		\label{fig:design4}
	} 
	\caption{System architectures for loading LLM. (a).The universal LLM loading method. (b).Compressed model size $<$ GPU memory capacity. There is a decompression buffer to store the decompression results. (c).GPU memory capacity $<$ Compressed model size $<$ GPU+CPU memory capacity. (d).GPU+CPU memory capacity $<$ Compressed model size. The decompression buffer is allocated in GPU memory to store the uncompressed data.}
	\label{fig:system architecture}
\end{figure*}

\noindent\textbf{System Architectures. }
The universal CPU-GPU architecture is shown in Figure \ref{fig:design1}. We assume that all inference computation processes are executed in GPU. The LLM model is initially stored in storage. For execution, the model data is loaded to CPU memory and then copied to GPU memory. During the model inference process, it reads weight data from GPU memory directly.

Depending on the memory size relative to the model size, there are three types of data transfer methods: 1) If the GPU memory can store the entire model data, this represents the fastest way to perform inference. All the necessary data is readily available in the GPU memory, allowing for efficient and rapid computations.
2) In cases where the GPU memory is insufficient to accommodate the entire model and the model is stored partially in the CPU memory, additional overhead is introduced due to swapping data between different memory locations (CPU and GPU). This swapping process can impact inference speed to some extent.
3) The slowest situation arises when both the GPU memory and available CPU memory are smaller than the size of the model. The weight data needs to be loaded from the storage into the GPU memory. However, accessing data from storage typically has a slower bandwidth compared to memory access. This bandwidth bottleneck significantly slows inference performance. 

\noindent\textbf{Inference Process.}
A compressed LLM model needs a decompression step in the inference process stage. 
Figures \ref{fig:design2}-\ref{fig:design4} illustrate our approach, where a decompression buffer is allocated to store the uncompressed weight data. The size of this buffer depends on the compression granularity of the model weight and can accommodate either a single tensor or multiple tensors.
If the GPU memory is sufficient for the compressed model and the small decompression buffer, all of the model data will be loaded into GPU memory, and the GPU core will be responsible for the model decompression and inference. For insufficient cases, it also includes a data swap with CPU memory, even with the storage. The compressed data can accelerate the transfer between different modules, especially when swapping with storage.

The GPU decompression speed is associated with the compressed size. 
Because of the GPU's decompression algorithm's parallel structure, a greater amount of data decompressed at once results in a faster decompression speed. The speed continues to increase until it reaches the maximum threshold based on the GPU's capabilities.
However, it's important to note that a larger buffer size also incurs additional memory overhead. Thus, finding an optimal balance between buffer size and decompression speed is crucial for efficient inference. 

\noindent\textbf{Throughput Analysis. }
The process of inference with a compressed model can be divided into three stages: data loading, decompression, and inference. During the data loading process, a chunk of compressed data is read and decompressed. This process includes three different types of loading, as shown in Figure \ref{fig:system architecture}. The bandwidth varies depending on the situation. The decompressed data is then stored in the decompression buffer in the GPU memory. In the inference stage, the weights are read from the decompression buffer, and the computations are performed. 

To analyze the inference speed, we consider factors such as the chunk size ($S_{chunk}$), number of chunks($N_{chunk}$), data access bandwidth($B_{loading}$), decompression speed($D_{gpu}$), and inference speed($I_{gpu}$). The data access bandwidth is decided by the bandwidth of storage-to-CPU($B_{stoc}$), CPU-to-GPU($B_{ctog}$), and GPU access($B_{gpu}$). The decompression speed refers to the rate at which the compressed weights can be decompressed per second, while the inference speed refers to the rate at which the weights can be computed per second. The per-sample model inference latency can be calculated using the following expression:

\begin{equation}
B_{loading} = \min(B_{stoc}, B_{ctog}, B_{gpu})
\label{equ:equation11}
\end{equation}
\begin{equation}
\text{L} = \max\left(\frac{S_{chunk}}{B_{loading}}, \frac{S_{chunk}}{D_{gpu}}, \frac{S_{chunk}}{I_{gpu}}\right)*N_{chunk}
\label{equ:equation12}
\end{equation}

\begin{figure}
    \centering
    \includegraphics[width=1\linewidth]{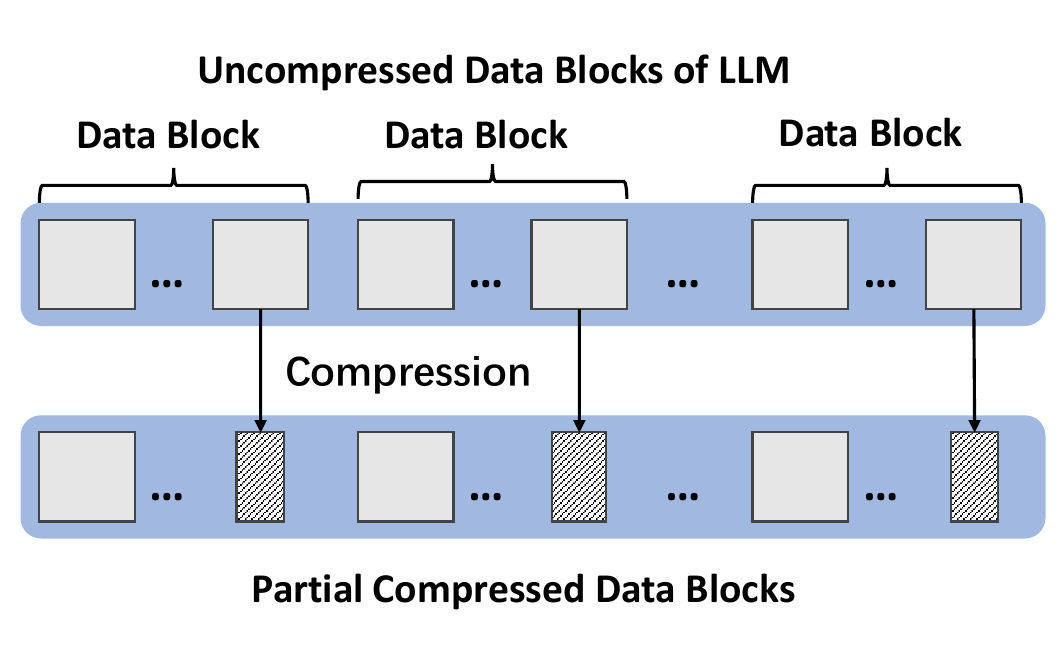}
    \caption{Speed Adaptive Compression. The LLM data is divided into data blocks containing several chunks. For every data block, the last chunk is selected for compression. The decompression speed can be increased by $Block_{size}/Chunk_{size}$ times at the cost of compression ratio loss.}
    \label{fig:adaptive compression}
\end{figure}

\noindent\textbf{Speed Adaptive Partial Compression. }
To address the bottleneck caused by decompression overhead and strike a balance between memory usage and inference speed, we present an adaptive partial compression for LLM. 
The weight data is divided into several granularity levels, such as tensor-wise or n-layer-wise, which we refer to as data chunks. These data chunks are then grouped together as data blocks.

As shown in Figure \ref{fig:adaptive compression}, all the uncompressed data in the LLM is divided into multiple data blocks, each containing N data chunks. 
We choose to compress only the last data chunk in each data block. This partial compression strategy allows for faster decompression during the inference process.

By compressing only a portion of the data, we trade off some of the model's compressibility. However, this trade-off enables us to reduce the memory requirement of LLMs while improving the decompression speed. The inference process now only needs to decompress the compressed data chunks, resulting in a decompression latency improvement of N times.

\section{Evaluations}

\subsection{Experimental Settings}
\noindent\textbf{Models and Datasets. }
We evaluate our methods DC in different LLM architectures: OPT-6.7b, Llama2-7b, Falocn-7b, and Mistral-7b models. 
The OPT models with six sizes (125m-13b) are evaluated to verify that DC is effective for different model sizes. 
We used a small calibration set from the Pile dataset to get the activation scale. 
The model performance is evaluated using Language Model Evaluation Harness\cite{eval-harness}. We select evaluation tasks from it: HellaSwag\cite{zellers2019hellaswag}, Lambada\cite{paperno2016lambada}, PIQA\cite{bisk2020piqa}, WinoGrande\cite{sakaguchi2021winogrande}, MMLU\cite{mmlu}, MathQA\cite{mathqa}, SWDE and Wikitext. 
The average accuracy is calculated to indicate the model performance.

\noindent\textbf{Baselines. }
We compare DC with FP16 and INT8 quantization.
The weight and activation are both quantized to INT8 for methods with quantization. 
We also provide aggressive results for per-channel scaling quantization and pruning schemes.
The Smoothquant\cite{smoothquant}, LLM.int8\cite{llmint8}, ZeroQuant\cite{zeroquant}, and OS+\cite{os+} quantization methods are compared with our quantization method.
The Magnitude\cite{magnitude}, Wanda\cite{wanda} and Pruner-Zero\cite{prunerzero} methods are compared with our pruning method.

\noindent\textbf{System Configuration.}
The decompression speed with different chunk sizes and the per-sample inference latency of models are evaluated in NVIDIA GPU A40 with 45GB memory. The compression/decompression interfaces are provided by NVCOMP library\cite{nvcomp}. We implement a high-performance decompression process into model inference on the GPU. 
The input data are DC-compressed LLMs.

\vspace{-0.1in}
\subsection{Compression Algorithms for LLMs}
To evaluate the different compression algorithms' performance for LLMs, we select five typical lossless algorithms and focus on the CR and Decompression Speed. The LZ4 and Snappy\cite{snappy} are dictionary-based algorithms. ANS is an entropy-based algorithm. Deflate\cite{deflate} combines Huffman\cite{huffman} and LZ77. Zstd\cite{Zstd} is supported by LZ77, Huffman, and ANS. All of them are implemented on GPU using nvcomp\cite{nvcomp} library.

The compression results are shown in Table~\ref{tab:algorithms}.
Deflate, ANS and Zstd show similar performance on model compression ratio. We can conclude that entropy-based algorithms are more suitable for LLM weight data. Because ANS only has one-step compression, the compression/decompression speed is much faster than others. So, we choose ANS as the compression algorithm for LLMs.
\begin{table}
  \centering
  \begin{tabular}{llll}
    \hline
     &\textbf{CR}& \textbf{CSpeed}&\textbf{DSpeed}\\
    \hline
    LZ4 &1.0137&1.88&19.07\\
    Snappy &1.0078&11.21&60.89\\
    Deflate &1.5000&1.50&7.91\\
    \textbf{ANS}&\textbf{1.4913}&\textbf{109.53}&\textbf{177.42}\\
    Zstd &1.4881&1.62&19.07\\
    \hline
  \end{tabular}
  \caption{Compression Algorithms Performance for OPT-1.3B-INT8. CSpeed is compression speed(GB/s) and DSpeed is decompression speed(GB/s). ANS is the most efficient compression algorithm for LLM compression.}
  \label{tab:algorithms}
\end{table}

\subsection{Model Compressibility}

\begin{table*}
    \centering
    \renewcommand{\arraystretch}{0.9}
    \resizebox{\linewidth}{!}{
    \begin{tabular}{ccccccccccl} \hline  
         &  &  &  \multicolumn{8}{c}{\textbf{Tasks (Accuracy$\uparrow$)}} \\   
         \textbf{Models}&  \textbf{Types}&  \textbf{CR}&  Hellaswag&  Lamabada&  PIQA&  Winogrande&  MMLU&  MathQA& SWDE &\textbf{AVG} \\ \hline
         & FP16& 0& 50.52& 67.65& 76.28& 65.59& 24.93& 24.66&  85.24&49.36\\
         OPT-6.7b& W8A8& 1.49& 50.34& 69.24& 76.22& 65.35& 24.98& 24.52&  85.78&49.55\\
         & DC-W8A8& 2.03 \textbf{($\uparrow$36\%)}& 49.72& 66.37& 74.86& 63.61& 25.47& 23.45&  84.88&48.55\\ \hline  
         &  FP16&  0&  57.31&  71.08&  78.07&  69.06&  41.77&  28.14&  87.67 &61.87\\  
         Llama2-7b&  W8A8&  1.84&  56.46&  68.84&  77.42&  69.14&  39.34&  29.48&  87.40 &61.15\\   
         &  DC-W8A8&  2.39 \textbf{($\uparrow$30\%)}&  56.46&  68.05&  77.91&  68.03&  38.62&  29.88&  87.31 &60.89\\ \hline  
         &  FP16&  0&  57.55&  71.56&  79.54&  67.25&  25.44&  28.78&  85.15 &59.32\\   
         Falcon-7b&  W8A8&  1.75&  57.30&  66.17&  78.73&  66.54&  26.59&  27.37&  82.72 &57.92\\   
         &  DC-W8A8&  2.00 \textbf{($\uparrow$14\%)}&  56.95&  67.33&  79.00&  66.22&  25.52&  27.44&  85.15 &58.23\\ \hline  
         &  FP16&  0&  61.2&  72.50&  80.74&  73.48&  58.61&  35.88&  90.19 &67.51\\   
         Mistral-7b&  W8A8&  1.95&  60.22&  70.80&  80.52&  72.95&  56.63&  33.91&  90.01 &66.43\\  
         &  DC-W8A8&  2.38 \textbf{($\uparrow$23\%)}&  60.55&  70.75&  80.03&  72.61&  56.44&  33.97&  90.12 &66.35\\ \hline 
    \end{tabular}}
    \caption{Model Performance. The W8A8 is supported by Smoothquant, which achieves the best CR in SOTA quantization methods.}
    \label{tab:model_performance}
\end{table*}

Our proposed method (DC-W8A8) offers a significantly higher post-quantization compression ratio (2.2x on average) while keeping comparable model performance (drop within 1\%) across various tasks, as illustrated in Table~\ref{tab:model_performance}. It improves the compressibility of the quantized models by 26\% on average and achieves above 2 times memory reduction compared with INT8 models. The Llama2-7b model can achieve a very high CR of up to 2.39, which means the INT8 quantized model can introduce an extra 58\% memory size reduction by our method. It's very promising to use DC to compress LLMs further.


\begin{figure}
    \centering
    \includegraphics[width=1\linewidth]{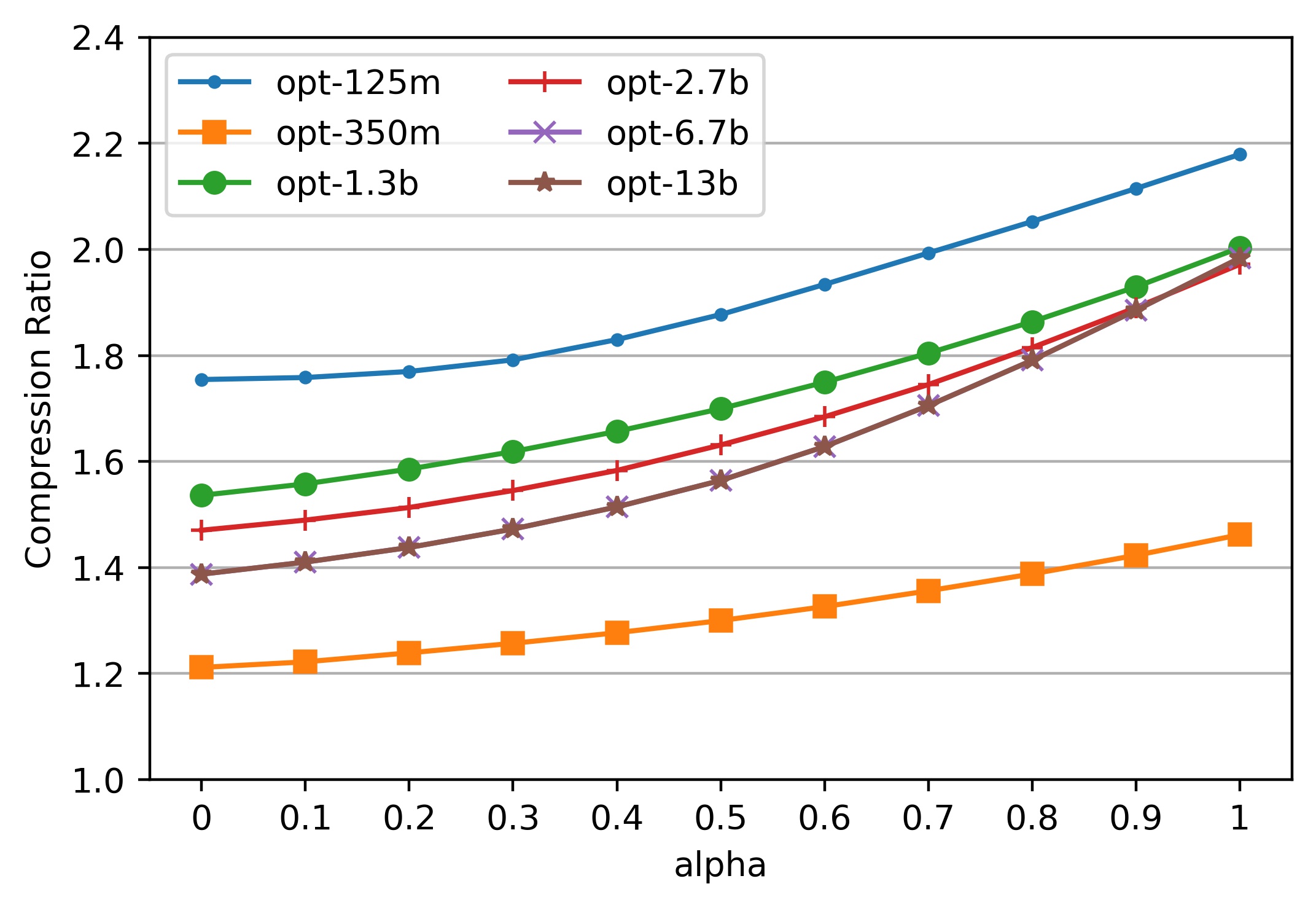}
    \caption{Model compression ratio with different scaling factors. Larger scaling factors can lead to better compression performance.}
    \label{fig: compression ratio}
\end{figure}

\subsection{Ablation Studies}

\noindent\textbf{Scaling Factors \& Model Sizes. }
Our objective is to achieve the best possible compression ratio while maintaining acceptable model accuracy. 
We conducted experiments on six OPT models with varying scaling factors ($\alpha$).  The compressibility improves as the scaling factor ($\alpha$) increases, as shown in Figure \ref{fig: compression ratio}. The majority of models typically achieve a compression ratio of approximately 2.0. 
By increasing the scaling factor, which introduces more outliers to the weight distribution, the compression ratios are enhanced by an average of 32.2\%. Particularly, on the opt-6.7B model, we achieved a remarkable reduction in model memory size, with a maximum compression ratio of up to 43\%.  It should be noted that the opt-350M has a lower compression ratio because the weight data distribution is more even than that of other models, even if we scale it with outliers.  

\begin{figure}
	\centering
	\subfigure[OPT-125M]{
		\begin{minipage}[b]{0.22\textwidth}
			\includegraphics[width=1\textwidth]{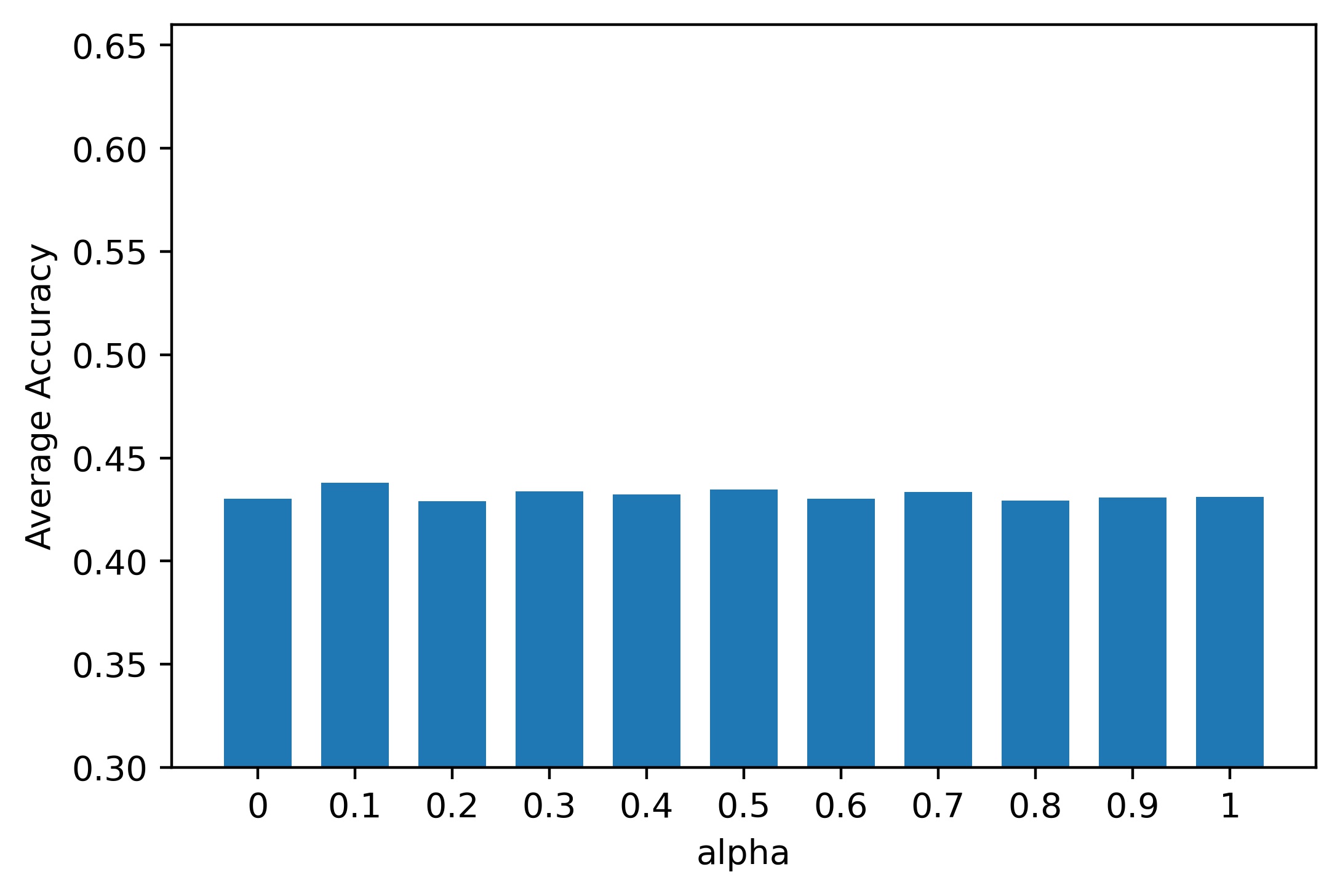}
		\end{minipage}
		\label{fig:Accuracy-125m}
	}
 	\subfigure[OPT-350M]{
		\begin{minipage}[b]{0.22\textwidth}
			\includegraphics[width=1\textwidth]{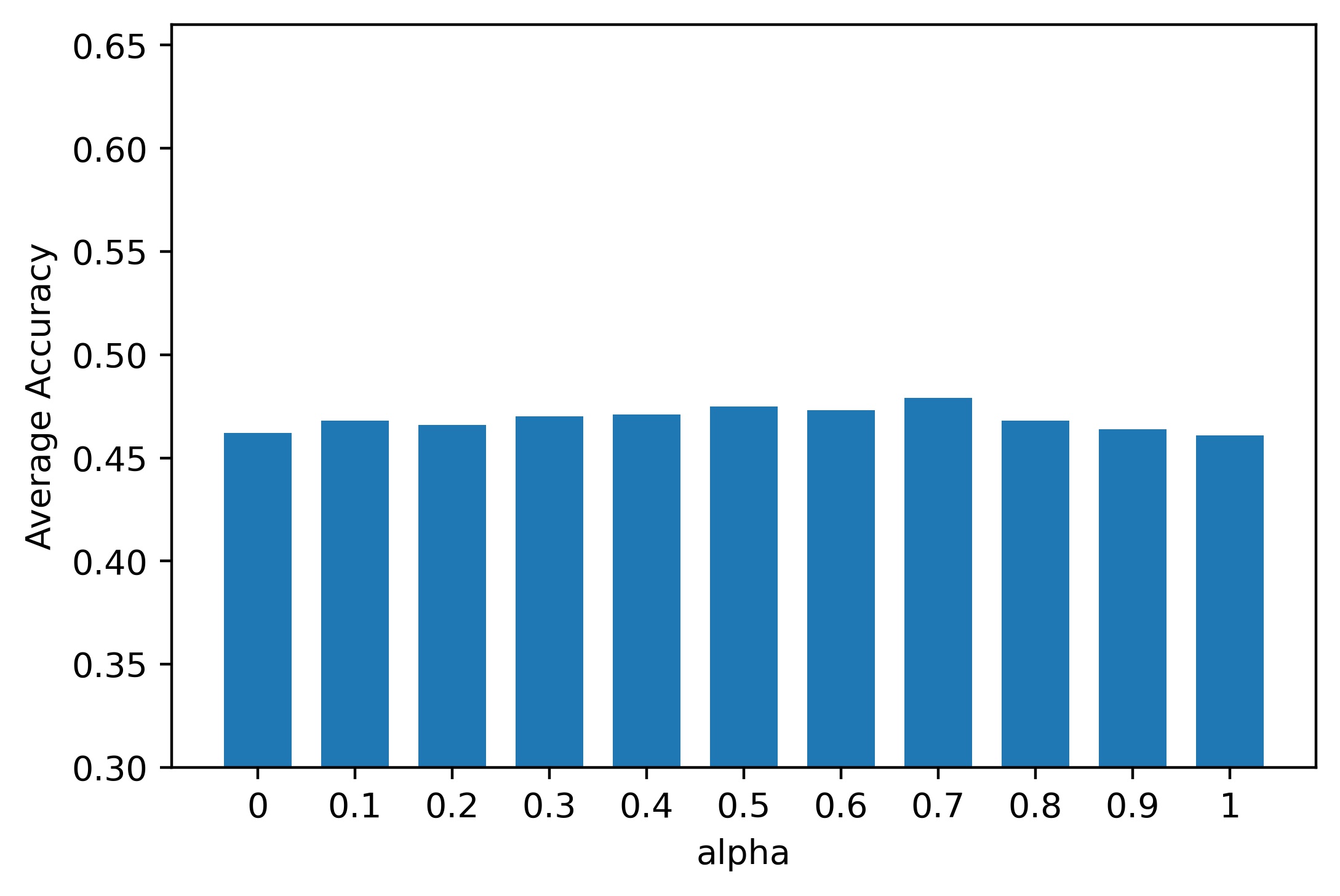}
		\end{minipage}
		\label{fig:Accuracy-350m}
	}
        \subfigure[OPT-1.3B]{
    	\begin{minipage}[b]{0.22\textwidth}
   		  \includegraphics[width=1\textwidth]{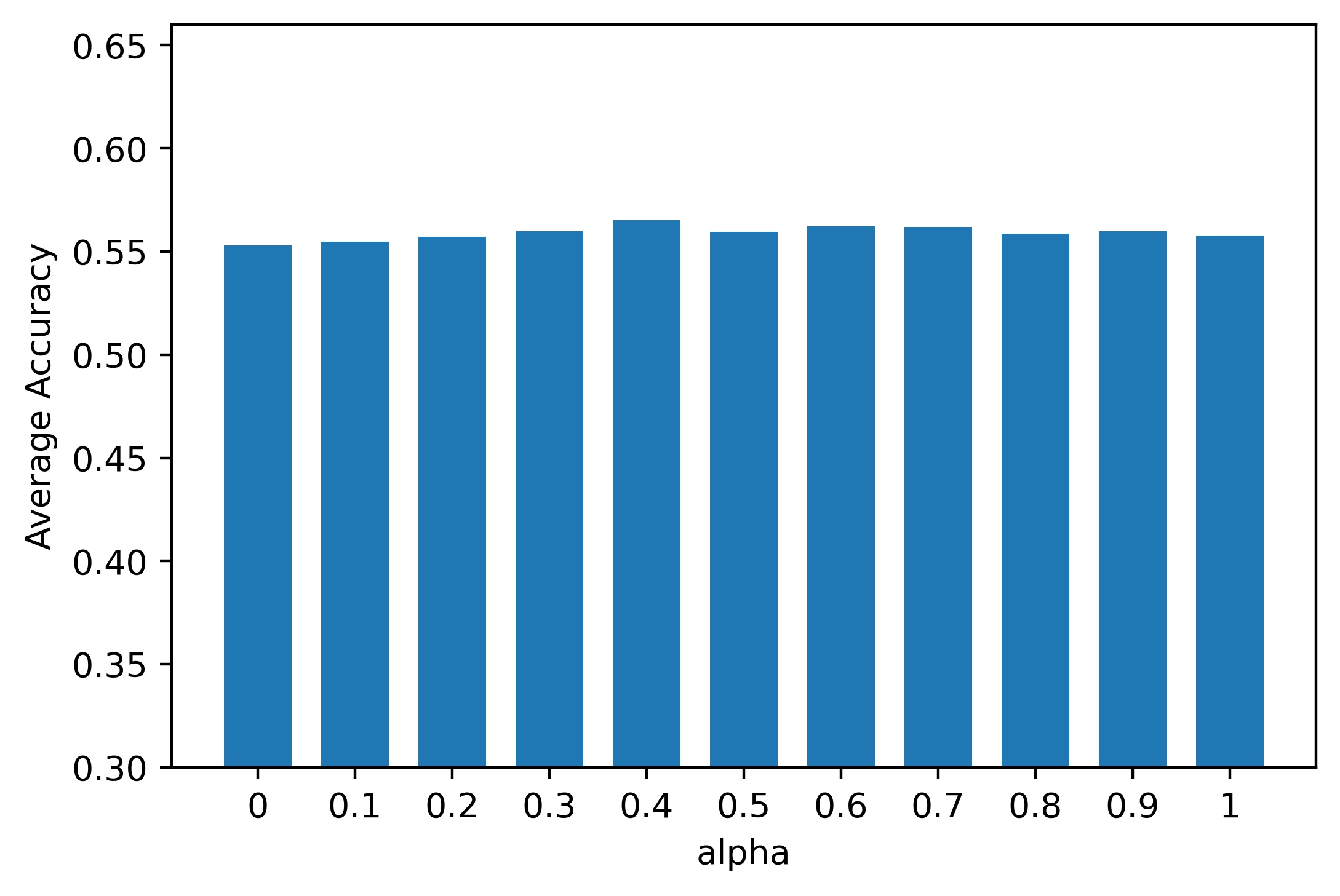}
    	\end{minipage}
		\label{fig:Accuracy-1.3b}
        }
        \subfigure[OPT-2.7B]{
		\begin{minipage}[b]{0.22\textwidth}
			\includegraphics[width=1\textwidth]{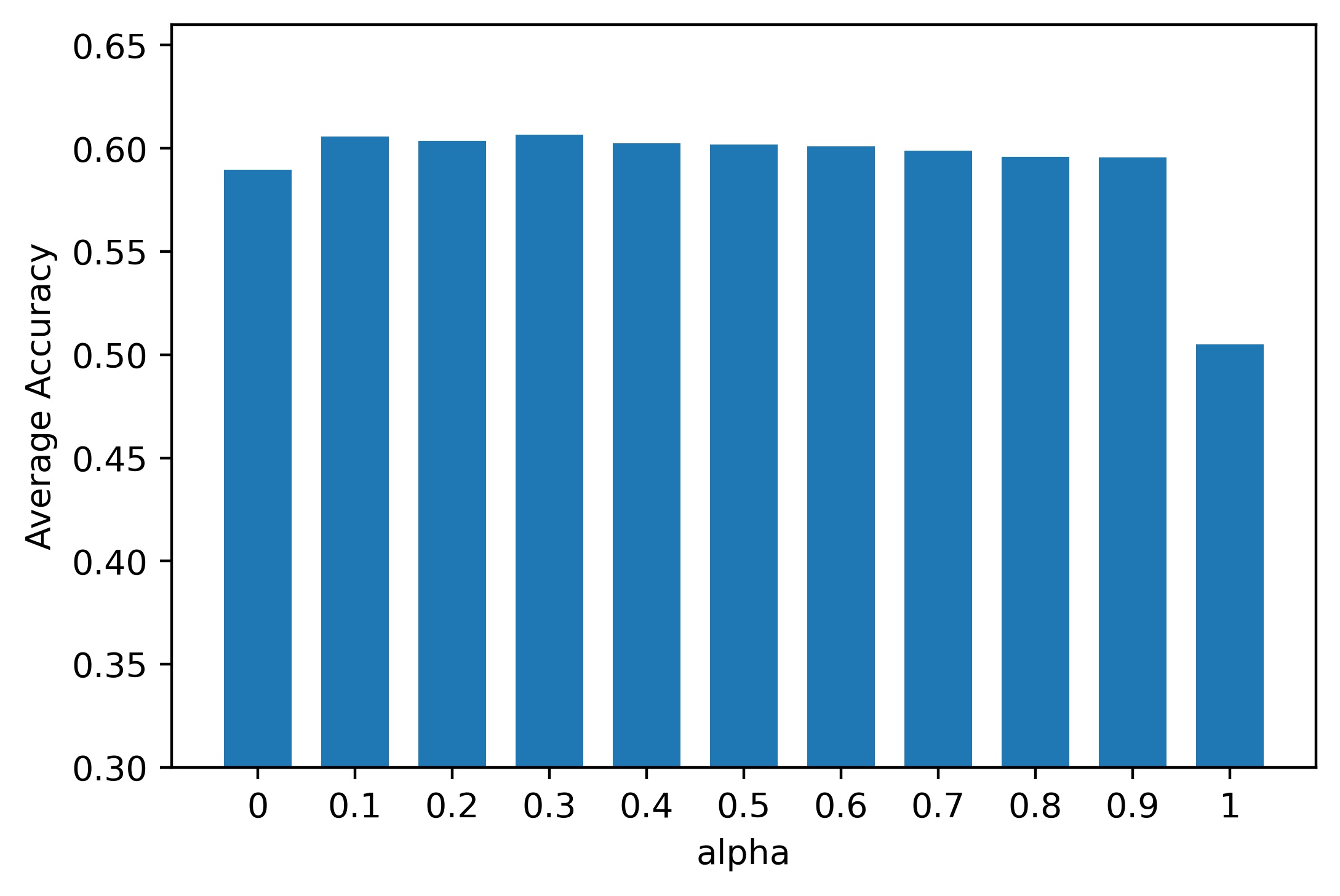}
		\end{minipage}
		\label{fig:Accuracy-2.7b}
	}
	\subfigure[OPT-6.7B]{
		\begin{minipage}[b]{0.22\textwidth}
			\includegraphics[width=1\textwidth]{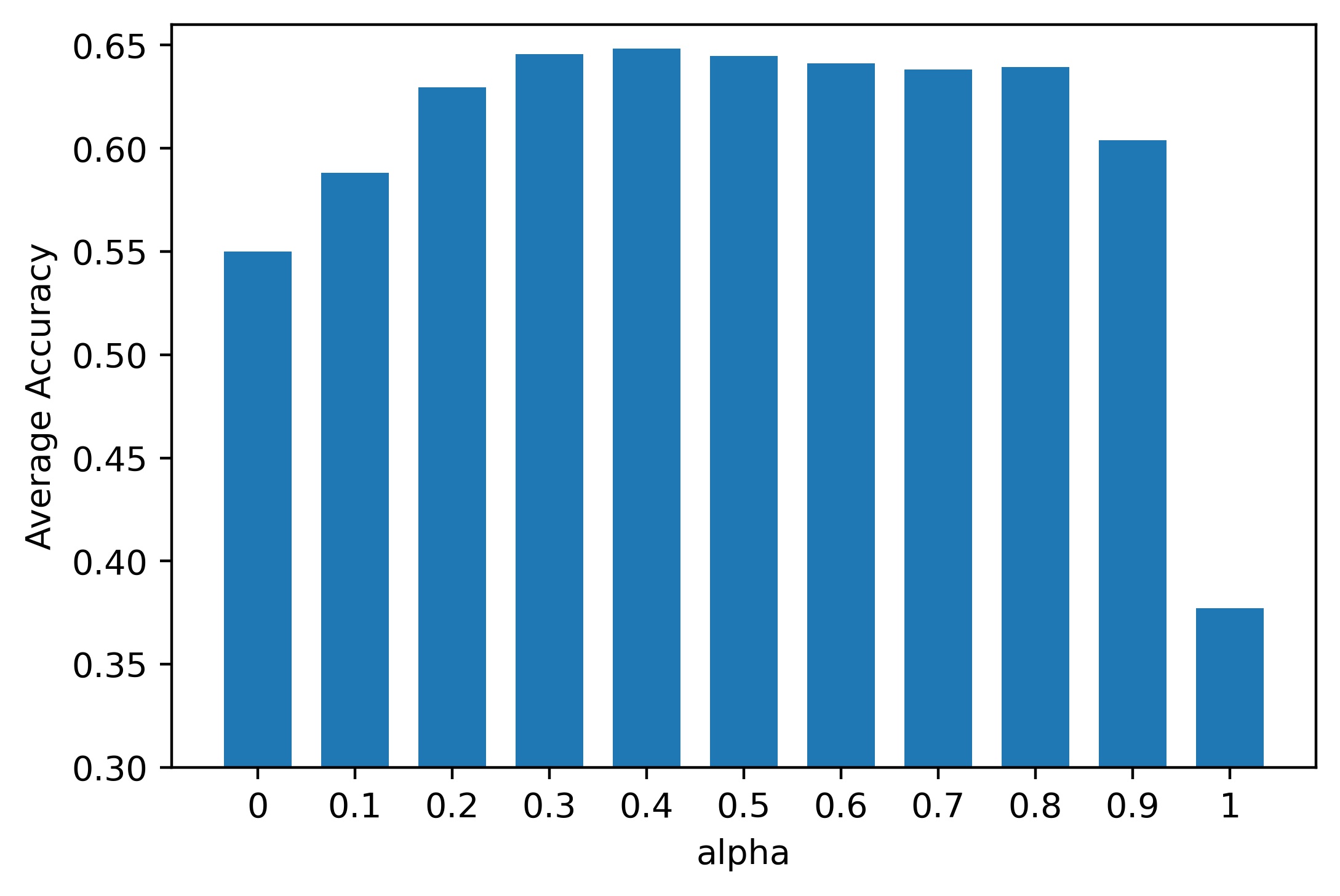}
		\end{minipage}
		\label{fig:Accuracy-6.7b}
	}
	\subfigure[OPT-13B]{
		\begin{minipage}[b]{0.22\textwidth}
			\includegraphics[width=1\textwidth]{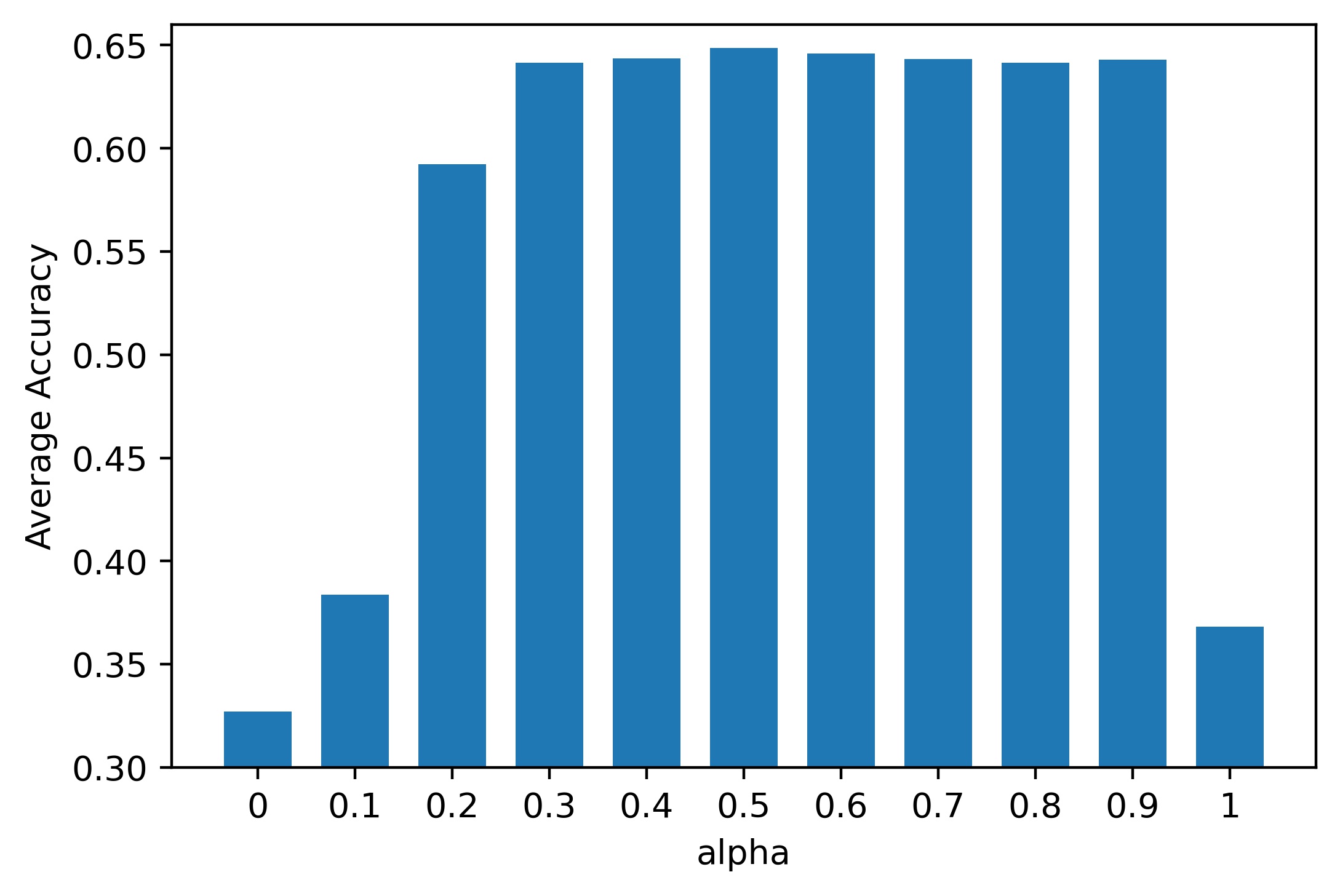}
		\end{minipage}
		\label{fig:Accuracy-13b}
	} 
	\caption{Average accuracy of double-compressed models is evaluated across different model sizes ranging from 125M to 13B.}
	\label{fig:acc for alpha}
\end{figure}

Figure \ref{fig:acc for alpha} shows the model accuracy with different scaling factors. The models with size below 1.3B show a small difference in accuracy with different $\alpha$. 
While opt-2.7B, opt-6.7B, and opt-13B show a sudden drop when the $\alpha$ changes from 0.9 to 1. The quantization error of weights is the reason here. With a small $\alpha$, the model accuracy still decreased due to the quantization error of activation. We can choose different scaling factors for different LLMs. 

\begin{figure}
    \centering
    \includegraphics[width=1.0\linewidth]{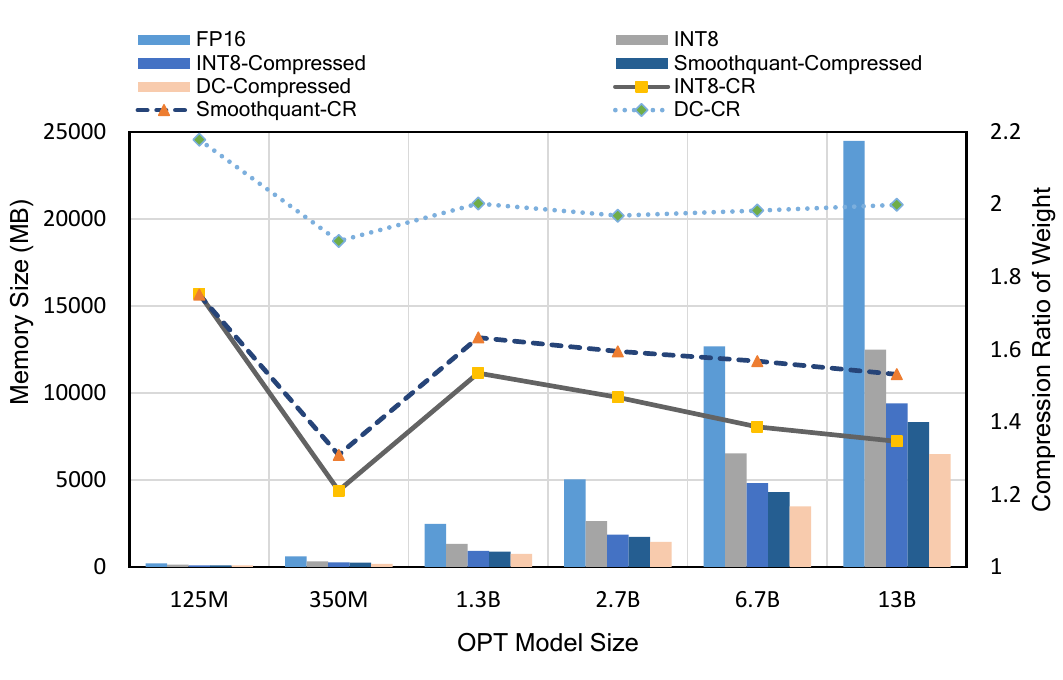}
    \caption{The comparison of memory size usage and compression ratio. Compared with smoothquant and INT8 quantization, our method exhibits better compression ratio and lower memory usage.
    }
    \label{fig:total improvement}
\end{figure}

\noindent\textbf{Memory Usage Reduction. }
We evaluate the memory usage for different types of models during the inference process. The results are shown in \ref{fig:total improvement}.
INT8 quantized data can reduce about 45\% memory usage reduction compared with the FP16 model, while INT8-compressed model 58\%. The further compression is an efficient method for LLMs to save more memory. Our DC-compressed LLM can achieve 67\% memory reduction, improved by 9\% and 7\% compared with INT8-Compressed and Smoothquant-Compressed methods.

\begin{figure}
    \centering
    \includegraphics[width=1\linewidth]{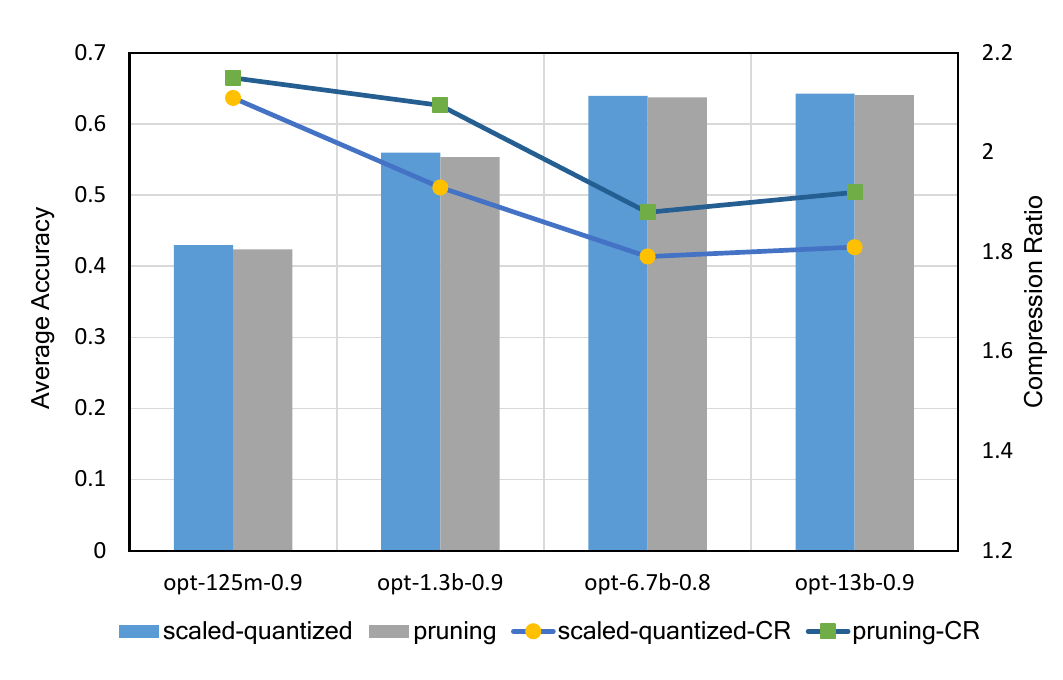}
    \caption{Average accuracy and compression ratio of quantized/pruned models with different schemes. The CR increases by 8.5\% for pruning while the loss of accuracy is negligible.}
    \label{fig:pruning result}
\end{figure}

\noindent\textbf{Aggressive results. }
We select models with four typical sizes and compare the pruned model with scaled-quantized models without pruning. We set the sparsity level at 20\%, indicating that about 20\% of the weight values will be set to zero. The sparsity parameters can be set higher to get a better CR while the model performance is acceptable.
Figure \ref{fig:pruning result} illustrates the model's compressibility and accuracy. The largest improvement in CR, up to 8.5\%, is observed in the quantized opt-1.3B model with a scaling factor of 0.9. Conversely, the quantized opt-125m model shows a smaller improvement of 1.9\%. This discrepancy can be attributed to the fact that the compression ratio of the opt-125m model is already quite high compared to the other models, leaving limited room for further improvement.

\noindent\textbf{Few-shot Tasks. }
For few-shot tasks, we compare the FP16 and DC-compressed Llama2-7b model using the Winogrande dataset. The results are shown in Table 2. The experiment result shows that our method does not impact model performance regardless of the number of shots.

\begin{table}
    \centering
    \resizebox{\linewidth}{!}{
    \begin{tabular}{lcccc} \hline 
         Model&  0-shot&  4-shot&  8-shot&  16-shot\\ \hline 
         Llama2-7b&  0.6906&  0.7482&  0.7419&  0.7506\\ 
         Llama2-7b-DC&  0.6803&  0.7245&  0.753&  0.7522\\ \hline
    \end{tabular}}
    \caption{Model performance for few-shot tasks using Winogrande dataset.}
    \label{tab:my_label}
\end{table}

\subsection{Comparative Evaluations}
\noindent\textbf{Quantization. }
Compared with different INT8 quantization methods, DC can achieve the best compressibility for quantized LLM. 
Compared with channel-wise and group-wise quantization, the tensor-wise quantized model has the best compressibility, which is the result of uneven data distribution. With the scaling method, the compression ratio can be further improved.  
Compared with Smoothquant, our method can improve the CR by 28\%.

\begin{table}   
    \label{tab:quant}
    \centering
    \resizebox{\linewidth}{!}{
    \begin{tabular}{|l|l|c|c|} \hline 
           Methods&Type&CR&  Wikitext PPL\\ \hline 
          Zeroquant&group-wise&1.15&  16.43\\
 LLM.int8()&per-channel& 1.21&16.56\\
 Baseline& per-tensor& 1.35&18.32\\   
          Smoothquant &scaled-tensor&1.57&  19.50\\ 
          Ours &scaled-tensor&2.13&  19.89\\ \hline 
    \end{tabular}}
    \caption{ Different Quantization Methods. Type is the quantization granularity.}
\end{table}

\noindent\textbf{Pruning. }
Compared with SOTA pruning methods: Magnitude\cite{magnitude}, Wanda\cite{wanda} and Pruner-Zero\cite{prunerzero}, our method shows better model performance. We evaluated these pruning methods on the quantized opt-1.3b model with the Wikitext dataset. The sparsity is set to 40\%. As shown in Table 2, Magnitude ignores the importance of activation for weight pruning and shows the worst PPL. Our method prunes the quantized weight with the same activation value, which gets the best PPL. Our method is also more time-efficient than Wanda and Pruner-Zero.

\begin{table}    
    \label{tab:pruning}
    \centering
    \resizebox{\linewidth}{!}{
    \begin{tabular}{|l|c|c|} \hline 
           Pruning Methods&Sparsity&  Wikitext PPL\\ \hline 
          Magnitude &40\%&  96.20\\   
          Wanda &40\%&  20.22\\ Pruner-Zero&40\%&  21.25\\ 
          Ours&40\%&  19.89\\ \hline 
    \end{tabular}}
    \caption{Different Pruning Methods on Quantized Model. }
\end{table}

\vspace{-0.1in}

\subsection{Inference with Compressed LLMs}

\textbf{Inference overhead.}
The overhead of inference with compressed LLMs can be analyzed on memory and computation.
The memory overhead is an additional decompression buffer, decided by the compression granularity, which is much smaller than the model size. The computation overhead consists of the decompression and Pytorch copy operation (convert the memory data into Pytorch tensors).
The evaluations are shown in Table ~\ref{tab:decompession speed}. Weight access speed is slower than computation speed, which will slow down the inference speed.

\begin{table}
  \centering
  \renewcommand{\arraystretch}{0.9}
  \begin{tabular}{lllll}
    \hline
     Size&Per layer& DSpeed&ASpeed &CSpeed\\
    \hline
    1.3B&48.05&97.76&25.41 &30.78\\
    2.7B&75.08&109.64&35.51 &45.03\\
    6.7B&192.13&144.01&70.17 &84.06\\
    13B&300.16&156.08&91.89 &105.10\\
    \hline
  \end{tabular}
  \caption{Decompression speed for different chunk sizes(MB). DSpeed is decompression speed(GB/s), ASpeed is weight data access speed(GB/s) with overhead and CSpeed is weight computation speed(GB/s). }
  \label{tab:decompession speed}
\end{table}


\begin{figure}
    \centering
    \includegraphics[width=1\linewidth]{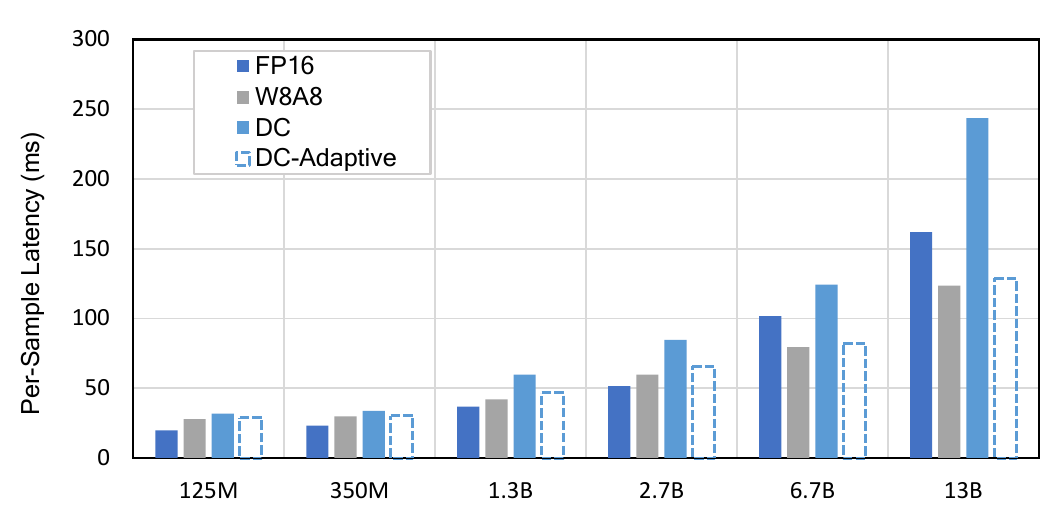}
    \caption{Per-sample inference latency of models with different quantization and compression methods. }
    \label{fig:per_sample_latency}
\end{figure}

\noindent\textbf{Inference Latency.}
The inference per-sample latency is evaluated to investigate the decompression overhead for LLM inference. As shown in Figure \ref{fig:per_sample_latency}, the latency increases by 32.86\% in average due to decompression overhead.
Applying the speed adaptive method, we select 1/5 of the model data to be compressed, which reduces memory size by 40\%. The latency can be reduced by 22.5\% and matched to the inference speed without decompression.
It's notable that the decompression and inference speed vary from device to device. We can trade off them by adaptive compressing the model.

\section{Related Works}
LLM compression\cite{wang2024modelsurvey}\cite{zhu2023survey} devotes to reducing the memory and computational cost of the model in the inference process. Quantization and Pruning are two typical compression methods for models\cite{deepcompression}.

Quantization reduces model size by converting floating point numbers to low-precision integers or other discrete forms. \cite{gptq} \cite{awq}\cite{llmint8} focus on weight quantization, while\cite{smoothquant} \cite{zeroquant} \cite{os+} focus on weight and activation quantization. These quantization works research how to reduce quantization error, and the compression ratio is fixed. 



Pruning reduces model parameter size by removing unnecessary or redundant components. \cite{llmpruner}\cite{gum} proposes structured pruning methods, which remove units based on specific rules. \cite{sparsegpt}\cite{wanda}\cite{prunerzero} aims at unstructured pruning methods. 

In this paper, we propose to compress quantized/pruned LLM further.


\section{Conclusion}
We propose a novel double compression method for LLMs, which combines model compression with lossless compression algorithms. Motivated by the observation that weight data in LLMs exhibit compressibility and the potential for further improvement in compression ratio, we present several approaches to enhance the compressibility. We also investigate the decompression overhead of inference with compressed models and propose an adaptive method to overcome the overhead. Our work provides a solution to reduce memory size and facilitate the deployment of LLMs.

\clearpage

\section{Limitations}

This work's limitation mainly lies in the quantization methods for LLMs. We focus on compressing INT8 quantized weights, as INT8 quantization is the most widely used method and is supported by lots of hardware. The model compressibility of quantized models may vary with different methods.  

\bibliography{acl_latex}

\end{document}